\documentclass{article}
\usepackage{graphicx}
\usepackage{wrapfig}
\usepackage{amsmath}
\usepackage{listings}
\usepackage[final]{corl_2026} % Uncomment for the camera-ready ``final'' version.

\title{MAGMA-GEN: Validated Recovery Supervision from Ambiguous Failures via Counterfactual Re-Execution}

\author{
  Loan Bernat\\
  Siléane, LAAS-CNRS \\
  \texttt{l.bernat@sileane.com} \\
  \And
  Matthieu Grard \\
  Siléane \\
  \texttt{m.grard@sileane.com} \\
  \AND
  Ariane Herbulot \\
  LAAS-CNRS \\
  \texttt{ariane.herbulot@laas.fr} \\
  \And
  Florent Lamiraux \\
  LAAS-CNRS \\
  \texttt{florent.lamiraux@laas.fr} \\
}

\begin{document}
\maketitle

%===============================================================================

\begin{abstract}
Hierarchical robotic systems executing long-horizon manipulation tasks must make high-level semantic decisions that orchestrate stochastic low-level skills. In this setting, failed rollouts are \textbf{ambiguous}: \emph{a poor downstream state may reflect an invalid high-level decision, partial observation, or a valid decision whose physical execution failed}. Traditional supervised learning lacks data for such recovery states, while reinforcement learning struggles with sparse rewards and non-local credit assignment. We propose \emph{MAGMA-GEN}, an on-policy data-generation pipeline that converts ambiguous failed rollouts into validated recovery supervision. MAGMA-GEN first uses a privileged coach to hypothesize an early decision-level error and propose localized correction or recovery actions. Because this diagnosis is fallible, candidates are retained only if re-execution from the same state under matched conditions improves downstream progress. This produces supervised examples from the agent's own failure distribution without per-step human demonstrations. Evaluated on interactive long-horizon manipulation tasks, \emph{MAGMA-GEN} achieves higher observed task-success and recovery rates than distillation and trajectory-repair baselines in simulation, while real-robot trials demonstrate transfer feasibility under evolving task constraints.\footnote{Project and code: \url{https://magma-rob.github.io/magma-gen}}
\end{abstract}

\keywords{Data Generation, Long-Horizon Planning} 

%===============================================================================

\section{Introduction}

Long-horizon interactive manipulation requires robotic systems to operate under partial observations and evolving rules where small errors compound over time. Unlike static, fixed-goal benchmarks \cite{Mees2021CALVINAB, PARTNR, zhang2024vlabench}, real-world deployment demands that agents adapt to dynamic user preferences, exceptions, and sequentially revealed instructions. Crucially, while agents may appear competent when trained and evaluated on isolated single-step instructions, they rapidly degrade in multi-turn interactive deployments \cite{laban2026llms}. Consequently, expert demonstrations cover only a narrow slice of the state space. Robust agents must instead learn from off-nominal states induced by their own planning mismatches or stochastic physical execution. While recent recovery-oriented frameworks \cite{rac-hu, Huang-2025-fail2progress} highlight the value of learning from failure, they typically focus on low-level visuomotor adjustments or rely heavily on external, human-provided supervision.

The key difficulty in exposing agents to failures during training, without human supervision, is that failed rollouts are \textbf{ambiguous}: a poor downstream state may result from an invalid high-level decision, a valid decision whose physical execution failed, or partial observation. Treating every failure as a decision error corrupts credit assignment, while ignoring failures prevents learning recovery behaviors. Effective learning therefore requires a way to disambiguate which failures should become decision-level supervision and which should instead induce recovery behavior. Existing reinforcement learning \cite{Feng2025ReToolRL, dalal2024planseqlearn} and trajectory-repair methods \cite{cleaner2026, autotraj, song-etal-2024-trial} only partially address this setting: they lack mechanisms to combine localized diagnosis with matched counterfactual validation under partial observability and action noise.

\begin{figure}[t]
    \centering
    \includegraphics[width=1\linewidth]{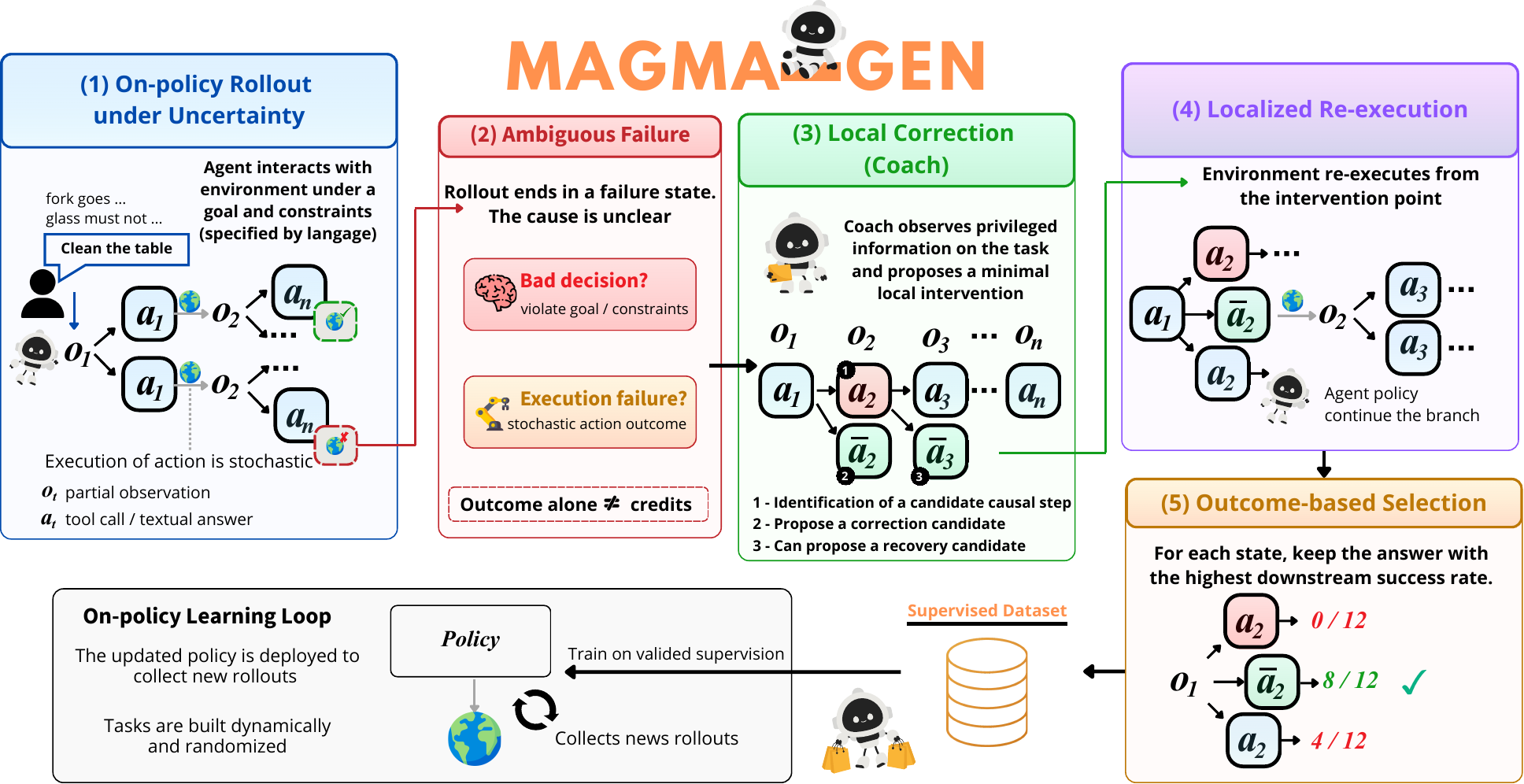}
    \caption{Overview of MAGMA-GEN. It converts ambiguous failed rollouts into supervision by proposing minimal local counterfactual interventions and retaining only those whose re-execution improves downstream progress.}
    \label{fig:magma_overview}
\end{figure}

We propose \textbf{MAGMA-GEN}, a failure-driven data-generation framework that addresses the scarcity of long-horizon recovery supervision by converting ambiguous on-policy failures into supervised training data for high-level robotic decision policies (Figure~\ref{fig:magma_overview}). Given a failed rollout, MAGMA-GEN uses a privileged coach to hypothesize an early decision-level error and propose localized counterfactual interventions, including both corrections and recovery actions. These candidates are then validated through simulator re-execution from the same state under matched conditions, retaining only interventions that improve downstream progress as supervision. Thus, MAGMA-GEN aims to disambiguate failures for training purposes without treating the coach diagnosis itself as ground truth. We evaluate MAGMA-GEN on interactive long-horizon manipulation tasks and find that it most consistently improves recovery from execution-induced failures and progress before failure.

%===============================================================================

\section{Related Work}
\label{sec:related-work}

\paragraph{Long-horizon robotic learning under interactive goals.} Language-conditioned robot learning spans end-to-end vision-language-action policies \cite{rt2, pmlr-v202-driess23a, Black-pi0}, modular planners \cite{Ahn2022DoAI, Liang2022CodeAP, Singh2022ProgPromptGS, Song2022LLMPlannerFG, dalal2024planseqlearn}, and more recently Tool-Integrated Reasoning (TIR) systems \cite{Paranjape2023ARTAM, Gou2023ToRAAT}. Benchmarks such as ALFRED, CALVIN, VLABench, and PARTNR emphasize multi-step embodied reasoning \cite{Shridhar2019ALFREDAB, Mees2021CALVINAB, zhang2024vlabench, PARTNR}. Unlike these settings, we study interactive tasks whose rules are updated during execution under partial observations, which makes expert data collection especially difficult.

\paragraph{Learning from failures and self-generated data.} TIR agents are often trained by supervised fine-tuning followed by reinforcement learning \cite{DeepSeekR1_2025, shang_rstar2}, a paradigm that remains difficult in long-horizon settings with sparse rewards, especially with small models \cite{shao2025spurious, yue2025limit-of-rlvr}. Inspired by dataset aggregation approaches such as DAgger \cite{dagger}, recent work generates synthetic supervision through reflection \cite{tool-llm-reflection, meta-early-exp}, exploration \cite{song-etal-2024-trial, Feng2025ReToolRL}, or self-play curricula \cite{languageselfplay, guidedselfevolvingllms, xia2025agent0}, but it typically does so over shorter horizons or requires expert data. More recent approaches \cite{cleaner2026, autotraj} extend this idea to multi-step trajectories. CLEANER \cite{cleaner2026} removes erroneous segments via rollback and AutoTraj \cite{autotraj} replaces failed trajectories with corrected ones. In contrast, MAGMA-GEN keeps failed on-policy rollouts and adds localized validated corrections, exposing the learner to failure states and recovery behavior within its own state distribution, while staying in a supervised learning objective.

\paragraph{Credit assignment under sparse or ambiguous feedback.} A central challenge in long-horizon decision-making is step-level credit assignment under sparse rewards. Hindsight-based \cite{hindsight-ca} and attribution-based \cite{pinpointing-ca} methods estimate per-step credit from trajectory outcomes, typically to update policies via reinforcement learning. By contrast, we use offline counterfactual re-execution to test whether a locally diagnosed intervention improves downstream viability, then train by supervised trajectory selection rather than reward attribution. Foundation models also support online failure handling: Sinha et al.~\cite{Sinha2024RealTime} combine anomaly detection with LLM-based fallback selection, while Li et al.~\cite{Li2026Hierarchical} combine VLM planning with reinforcement-learned skills and failure recovery for deformable-object routing. MAGMA-GEN instead uses failure reasoning during data generation to train a policy deployed without a coach or simulator. Execution-based validation matters in robotics because failures do not reliably reflect decision quality: stochastic execution and partial observability can make valid decisions look bad, and invalid ones look locally benign. MAGMA-GEN uses the simulator as a matched re-execution engine, allowing proposed corrections and recoveries to be filtered by their observed downstream effect.

%===============================================================================

\section{Problem formulation}
\label{formulation}

\begin{figure}[t]
    \centering
    \includegraphics[width=1\linewidth]{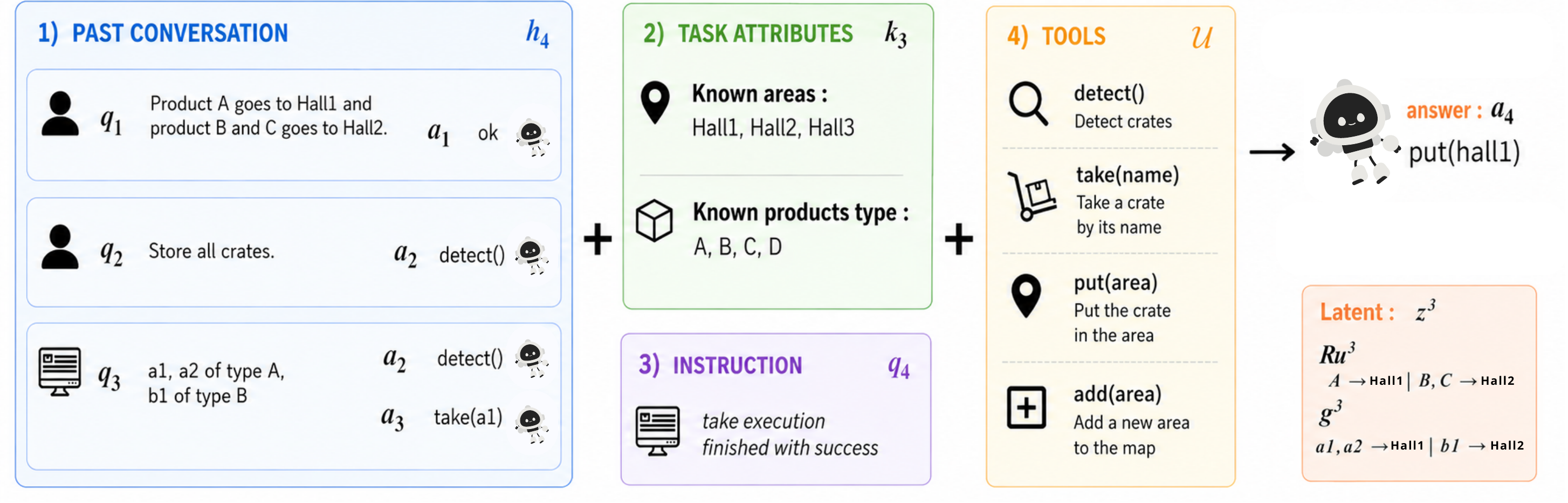}
    \caption{Example task in the MAGMA-GEN setting. At time $t$, the instruction is partially specified and relies on rules introduced earlier in the interaction. These rules can be overridden during execution, creating long-horizon settings where decision errors and execution failures compound over time.}
    \label{fig:task}
\end{figure}

Unlike fully specified manipulation benchmarks \cite{PARTNR, Mees2021CALVINAB, zhang2024vlabench}, we consider long-horizon language-conditioned robot tasks in which goals and rules are introduced through interaction, as illustrated in Fig.~\ref{fig:task}. We focus on \textbf{task-level decision making}: the agent produces textual responses and selects stochastic low-level robotic skills such as \textit{take(name)}, \textit{pour\_in(name)}, or \textit{close\_door()}. These low-level skills, referred to as tools, may be implemented as 3D policies \cite{Garcia2024TowardsGV}, visuomotor policies \cite{act-RSS-23}, or motion planners \cite{hpp}. Unlike generic software tool-use, each tool call triggers a stateful robot skill whose outcome depends on physical state, perception, reachability, and stochastic execution. Therefore, a valid high-level decision can still produce an off-nominal physical state requiring recovery.

\subsection{Task state}
\label{task-formulation}

A task is represented as $\mathcal{T}=(\mathcal{G},\mathcal{R}_{\mathrm{u}},\mathcal{K},\mathcal{U})$, where $\mathcal{G}$ is the goal space, $\mathcal{R}_{\mathrm{u}}$ the space of interaction-derived task rules, $\mathcal{K}$ the structured task attributes available to the agent, and $\mathcal{U}=\{u_i\}_{i=1}^{M}$ the available tools. Each tool $u_i$ has an argument space $\mathcal{X}_i$.

At time $t$, the active goal $g^t \in \mathcal{G}$ and active rules $R_{\mathrm{u}}^t \subseteq \mathcal{R}_{\mathrm{u}}$ define a latent task component
\begin{equation}
\label{eq:task-state}
    z^t = (g^t, R_{\mathrm{u}}^t).
\end{equation}

\subsection{POMDP formulation}

We model execution as a factored partially observable Markov decision process (POMDP) \cite{ReinforcementLA,Kaelbling1998POMDP}
\begin{equation}
    \mathcal{P} = \langle \mathcal{S}, \mathcal{A}, \mathcal{O}, T, \Omega, \rho_0 \rangle.
\end{equation}
Here, $\mathcal{A}$ contains natural-language responses paired with either no tool call or a tool invocation, $\mathcal{O}=\mathcal{K}\times(\mathcal{I}\cup\mathcal{E})$, and $T$, $\Omega$, and $\rho_0$ are the transition, observation/event, and initial-state kernels. With latent physical world state $w^t$ and task attributes $k^t$, the full state is $s^t = (w^t, z^t, k^t) \in \mathcal{S}$. We do not assume access to a per-step reward, but only a sparse binary task-completion signal. The agent outputs a structured action
\begin{equation}
    a^t = (a_u^t, a_m^t) \in \mathcal{A},
\end{equation}
where $a_u^t$ is a natural-language response and $a_m^t=(u_i^t, x_i^t)$ a tool invocation with arguments $x_i^t \in \mathcal{X}_{i^t}$.
The agent does not observe $s^t$ directly. Instead, it receives
\begin{equation}
    o^t = (k^t, q^t) \in \mathcal{O},
\end{equation}
where $q^t$ is either a user instruction $\in \mathcal{I}$ or feedback from the preceding tool execution $\in \mathcal{E}$. Instructions revise the latent task specification, whereas tool feedback reports physical execution outcomes and may be stochastic or incomplete. We define the interaction history before the current observation as $h^t = (o^1, a^1, \dots, o^{t-1}, a^{t-1})$. To remain agnostic to the agent architecture, we let $m^t = f(h^t)$ denote the agent's internal representation of this history, where $f$ may include operations such as context truncation, summarization, memory retrieval, or latent-state updates. We finally write the policy as
\begin{equation}
    \pi_{\theta}(a^t \mid m^t, o^t, \mathcal{U}),
\end{equation}
which models decision making as next-action prediction conditioned on the interaction history and the current observation. Details on the latent dynamics are provided in Appendix~\ref{app:problem-formalization}.

\subsection{Validity and ambiguity under stochastic execution}
We distinguish decision correctness from execution outcome through a latent validity variable
\begin{equation}
\mathrm{valid}(a^t, z^t) \in \{0,1\},
\end{equation}
which indicates whether action $a^t$ is consistent under the latent task state $z^t$. An invalid action corresponds to a \emph{decision error}, while a valid action may still fail due to stochastic execution. This validity signal is never directly observable. After acting, the agent only receives an interaction outcome $q^{t+1} \in \mathcal{I} \cup \mathcal{E}$ generated by stochastic environment dynamics. As a result, observed failures are causally ambiguous: the same negative outcome may arise either from an invalid decision or from execution noise despite a valid decision. This ambiguity motivates mechanisms that propose localized credit assignments, but validate them through counterfactual execution rather than accepting outcome attribution directly.

\section{Method: Failure-Driven Supervision via Coaching}
\label{sec:coaching}
\label{sec:magma-gen}

\subsection{Overview}

MAGMA-GEN converts ambiguous failed on-policy rollouts into supervised training data. Starting from the policy $\pi_{\theta}(a^t \mid m^t, o^t, \mathcal{U})$ defined in Sec.~\ref{formulation}, we collect rollouts
\begin{equation}
    \tau = (o^1, a^1, \dots, o^H, a^H),
\end{equation}
under evolving goals and rules $z^t$ and stochastic world dynamics. Because failed outcomes are ambiguous with respect to decision validity, they cannot be used directly as supervision.

MAGMA-GEN uses a privileged coaching module as a counterfactual intervention proposer (see Figure~\ref{fig:magma_overview}). Given a failed robot rollout, the \textbf{coach} hypothesizes an early decision-level error and suggests a local correction or recovery action. This hypothesis is not used as a label by itself: the candidate intervention is re-executed from the same physical state and retained only if it improves downstream completion.

\subsection{Privileged local intervention}

Considering long-horizon tasks with evolving goals and rules, we divide each trajectory into stages. Each stage $j$ corresponds to an active goal $g^{(j)}$ and active rules $R_{\mathrm{u}}^{(j)}$. Let $\tau^{(j)}$ denote the subtrajectory executed while stage $j$ is active. When this subtrajectory fails to complete $g^{(j)}$ under $R_{\mathrm{u}}^{(j)}$, the coaching system performs failure typing, diagnosis, and counterfactual intervention.

\paragraph{Failure typing.}
The system first assigns a coarse failure category $c^{(j)}$ to restrict the intervention space based on the simulator state. The taxonomy and attribution heuristics are given in Appendix~\ref{app:failure}.

\paragraph{Localized diagnosis.}
Conceptually, the ideal intervention point is the first action in stage $j$ that violates the active task specification due to a decision error:
\begin{equation}
    \ell^\star = \min \{ t \mid \mathrm{valid}(a^t, z^t) = 0 \}.
\end{equation}
Since $\mathrm{valid}(a^t, z^t)$ is unobservable, the coach is modeled as a proposal distribution rather than an oracle. It samples a candidate intervention index $\ell$ and a structured diagnosis $d^{(j)}$ from the stage segment $\tau^{(j)}$, the failure type $c^{(j)}$ and the privileged active task specification $z^{(j)}$:
\begin{equation}
    (\ell, d^{(j)}) \sim
    q_{\phi}^{\mathrm{diag}}(\cdot \mid \tau^{(j)}, c^{(j)}, z^{(j)}).
\end{equation}
Here $\phi$ denotes the coach implementation. In our experiments, $q_{\phi}^{\mathrm{diag}}$ is instantiated with an LLM prompt built from these inputs and constrained to return the suspected decision-error step and a short diagnosis. The privileged variable $z^{(j)}$ is available only during MAGMA-GEN data generation, because training episodes are sampled from a structured task generator (Appendix~\ref{app:task-building}). The agent never observes it directly. The diagnosis is therefore a targeted proposal for where supervision may be useful, while re-execution determines whether the proposal is retained.

\paragraph{Counterfactual intervention.}
Conditioned on the diagnosis, the coach proposes a small set of candidate interventions
\begin{equation}
    \mathcal{I}_{\phi}^{(j)}
    \sim q_{\phi}^{\mathrm{corr}}(\cdot \mid d^{(j)}, a^{\ell}, z^{(j)}, \mathcal{U}),
    \qquad
    \mathcal{I}_{\phi}^{(j)} \subseteq \{\tilde{a}^{\ell}, \tilde{a}^{\ell+1}\}.
\end{equation}
In our implementation, $q_{\phi}^{\mathrm{corr}}$ is a second structured LLM prompt constrained to the same action and tool-call schema as the policy. A correction $\tilde{a}^{\ell}$ replaces the likely faulty action at $\ell$, whereas a recovery action $\tilde{a}^{\ell+1}$ is inserted after it to react from the resulting off-nominal state.

Both interventions operate within the existing high-level tool-call interface. A low-level execution failure need not itself be a decision error: after a valid grasp attempt fails, the policy may incorrectly issue a placement command as if it had succeeded. In this example, $\ell$ identifies the subsequent placement decision, and a correction can replace it with another admissible grasp attempt. MAGMA-GEN thus learns appropriate high-level responses to execution failures without modifying the underlying perception or control skills.

In this way, the coach provides targeted exploration around suspected decision errors and execution failures: instead of sampling many independent rollouts, MAGMA-GEN allocates additional counterfactual trials to the steps most likely to make the rollout non-viable. Matched re-execution then separates useful corrections or recoveries from misleading diagnoses.

\subsection{Rollout tree}

\paragraph{Re-execution.} Because coach proposals are fallible, each proposed intervention $\tilde{a}^{t}$ is treated as a candidate branch rather than as a label. We restore the simulator to the parent state $s^t$ at which $a^{t}$ was selected, execute $\tilde{a}^{t}$, and let the current policy continue until the stage succeeds or fails. The original rollout and its counterfactual continuations are stored as a tree, so the same parent state can have multiple outgoing answers: policy-sampled actions and coach-proposed interventions.

\paragraph{Branching Factor.}
We also increase the density of collected data by sampling a small action set from the agent policy, instead of a single action, at each expanded state:
\begin{equation}
    \mathcal{B}_\theta(m^t,o^t) = \{a_1^t,\dots,a_B^t\}, \qquad
    a_b^t \sim \pi_\theta(\cdot \mid m^t, o^t, \mathcal{U}),
\end{equation}
where each action defines a child branch. At diagnosed intervention points, coach interventions add targeted counterfactual branches to the same tree.

\paragraph{Structured Uncertainty.}
Trees are collected under the structured execution and observation uncertainty from Sec.~\ref{formulation}, including grasp instability, reachability failures, partial detections, and missing feedback. This exposes the learner to post-failure states induced by stochastic robot execution, not only by invalid high-level decisions. For fair branch comparisons, perturbations are sampled once at the beginning of the stage and held fixed across sibling branches. Appendix~\ref{app:exploration} details the expansion control and user-simulation kernel.

\subsection{From trajectories to supervised learning}

MAGMA-GEN converts the rollout tree into a supervised dataset by selecting, for each reached state, the action with the best validated downstream continuation. Let $\mathcal{A}(s^i)$ denote all actions evaluated from state $s^i$, regardless of their origin: policy-sampled answers and coach-proposed interventions are all treated as candidates in the same set. For any $a \in \mathcal{A}(s^i)$, let $\mathcal{R}_{K}(s^i,a)$ denote the $K$ descendant rollouts obtained by executing $a$ and then continuing with the current policy. The subtree-success score is the fraction of these continuations that complete stage $j$:
\begin{equation}
    \label{eq:subtree_success}
    \mathrm{Succ}_j(s^{i}, a)
    =
    \frac{1}{|\mathcal{R}_{K}(s^i,a)|}
    \sum_{\tau_r \in \mathcal{R}_{K}(s^i,a)}
    \mathbf{1}\{\tau_r \text{ completes stage } j\}.
\end{equation}
For each reached state $s^i$, we select the highest-scoring action
\begin{equation}
    \label{eq:suc}
    a^\star(m^i,o^i)
    =
    \arg\max_{a \in \mathcal{A}(s^i)}
    \mathrm{Succ}_j(s^i, a),
\end{equation}
and add $(m^i,o^i,a^{\star})$ to $\mathcal{D}_{\text{MAGMA}}$ only if $\mathrm{Succ}_j(s^i,a^\star)>0$. Thus, we only retain the best answer for each state without privileging coaching answers. States for which every available action has zero downstream success are discarded to avoid training on uninformative failures.

The policy is then updated with a standard supervised objective
\begin{equation}
    \mathcal{L}_{\text{MAGMA}}(\theta) =
    - \sum_{(m^t, o^t, a^t) \in \mathcal{D}_{\text{MAGMA}}}
    \log \pi_\theta(a^t \mid m^t, o^t, \mathcal{U}).
\end{equation}
Because retained branches originate from trajectories sampled from the current policy and are re-executed from matched parent states, the extracted supervision is grounded in the policy's own state distribution while covering recovery states absent from curated expert datasets.

%==================================

\section{Experiments}
\label{sec:result}

Our evaluation tests a specific claim: MAGMA-GEN improves the generation of
supervised recovery data from ambiguous robot failures. Accordingly,
we separate local recovery, long-horizon task completion, mechanism ablations,
and physical transfer.

\begin{itemize}
    \item \textbf{Q1: Recovery.} Under the same seeded task allocation, does
    validated counterfactual re-execution improve recovery on a fixed held-out
    suite with identical injected-error schedules across methods?
    \item \textbf{Q2: Long-horizon transfer.} Do local recovery gains translate
    into complete scenario success or progress before failure?
    \item \textbf{Q3: Scaling and robot transfer.} What is the effect of increasing the branching factor, and do the gains transfer to real-robot execution?
\end{itemize}

\subsection{Setup}

\paragraph{Protocol.}
All trained policies start from Qwen3-4B \cite{qwen3technicalreport}. In each of four generation runs, every method receives the same 180 seeded task instances, with identical ordered subgoals and target-step limits. Best-of-N and CLEANER are capped at $N=10$ retries or correction-and-relaunch attempts per subgoal. AutoTraj repairs failed 20B expert trajectories post hoc, whereas MAGMA-GEN uses a 4B rollout policy and a 20B coach to diagnose failed subtrajectories and propose localized interventions that are validated by re-execution. All collected datasets fine-tune the same Qwen3-4B policy. ManiSkill3 \cite{taomaniskill3} is the simulator. Full baseline definitions are in Appendix~\ref{app:protocol}.

\paragraph{Benchmark and metrics.}
The simulated benchmark contains 279 held-out tasks across four scenarios (Make Coffee, Warehouse Sorting, Laundry, and Color Sorting), with horizons ranging from 3 to 18 high-level actions. Physical-transfer runs are reported separately and are not included in the simulated metrics. We report cumulative success rate (CSR), completed goal fraction before the first unrecoverable failure (CGC), and recovery rate. Recovery is evaluated on a specific held-out recovery suite of 60 unique tasks: it is computed over a fixed set of perturbed subgoals and measures whether the agent completes the active subgoal despite injected errors such as missed detections, failed grasps, or unreachable-object feedback.

\subsection{Main Results}

\begin{table}[t]
\centering
\footnotesize
\caption{Simulated evaluation under the same 180 seeded task instances. Results are means$\pm$std over four generation runs.}
\label{tab:main_multiseed}
\resizebox{1\linewidth}{!}{%
\begin{tabular}{l|l|c|c|c|c|c}
\hline
\textbf{Method} & \textbf{Failure Handling} & \textbf{Rollout} & \textbf{Corrector}
& \textbf{Recovery} & \textbf{CSR} & \textbf{CGC} \\
\hline
Best-of-N Distill. & Retry & 4B & -- & 1.66{\scriptsize $\pm$ 1.44} 
& 4.99{\scriptsize $\pm$ 0.38} & 14.57{\scriptsize $\pm$ 1.57} \\
CLEANER-adapted~\cite{cleaner2026} & Correction & 4B & 20B & 2.23{\scriptsize $\pm$ 0.48}
& 2.4{\scriptsize $\pm$ 0.07} & 11.34{\scriptsize $\pm$ 0.57} \\
\hline
Expert Rollout Distill. & None & 20B & -- & 12.50{\scriptsize $\pm$ 1.07}
& 7.3{\scriptsize $\pm$ 1.07} & 14.06{\scriptsize $\pm$ 1.68} \\
CLEANER-adapted~\cite{cleaner2026} & Correction & 20B & 20B & 12.77{\scriptsize $\pm$ 2.08}
& 15.82{\scriptsize $\pm$ 1.90} & 28.24{\scriptsize $\pm$ 2.96} \\
AutoTraj-adapted~\cite{autotraj} & Post-hoc Trajectory Repair & 20B & 20B & 5.00{\scriptsize $\pm$ 2.50}
& 12.93{\scriptsize $\pm$ 1.77} & 23.60{\scriptsize $\pm$ 3.85} \\
Best-of-N Distill. & Retry & 20B & -- & 16.66{\scriptsize $\pm$ 1.36}
& 14.15{\scriptsize $\pm$ 2.76} & 26.73{\scriptsize $\pm$ 4.84} \\
\hline
\textbf{MAGMA-GEN ($B=1$)} & \textbf{Diagnose + Correction + Recovery} & 4B & 20B & \textbf{26.25}{\scriptsize $\pm$ 3.75}
& \textbf{16.79}{\scriptsize $\pm$ 2.98} & \textbf{30.74}{\scriptsize $\pm$ 3.98} \\
\hline
\end{tabular}%
}
\end{table}

\paragraph{Q1: Recovery from injected execution failures.}
Table~\ref{tab:main_multiseed} directly evaluates our primary claim. MAGMA-GEN achieves the highest mean recovery rate at \textbf{26.25\%}, a 9.59 percentage-point gain over the strongest baseline at \textbf{16.66\%}, despite using a 4B rollout policy rather than 20B expert rollouts. AutoTraj improves CSR and CGC over raw Expert Rollout Distillation but reduces recovery from 12.50\% to 5.00\%, indicating that repairing complete trajectories does not preferentially generate recovery supervision. CLEANER's immediate corrections improve long-horizon performance over raw expert rollouts, but reach only 12.77\% recovery, suggesting that correcting the detected failure without diagnosing the responsible decision yields lower-quality recovery supervision. Section~\ref{sec:ablation} tests this explanation by comparing correction validity and isolating the role of re-execution.

\paragraph{Q2: From local recovery to long-horizon progress.}
MAGMA-GEN also obtains the highest mean CSR (16.79\%) and CGC (30.74\%) in Table~\ref{tab:main_multiseed}. This joint improvement is consistent with recovery supervision helping the policy progress farther rather than trading recovery against long-horizon performance. Absolute CSR nevertheless remains low for every method, reflecting compounding errors over horizons of up to 18 high-level actions.

\begin{table}[h]
\centering
\footnotesize
\caption{Collection resources and data yield on the same 180 seeded task instances, averaged over four generation runs ($B=1$ where applicable). Counts are rounded to the nearest call, transition, or example.}
\label{tab:collection-budget}
\begin{tabular}{l|c|c|c|c}
\hline
\textbf{Resource / Yield} & \textbf{MAGMA-GEN} & \textbf{Best-of-N} & \textbf{CLEANER} & \textbf{AutoTraj} \\
\hline
20B calls & 3585 & 6538 & 6440 & 6313 \\
4B rollout calls & 5012 & 0 & 0 & 0 \\
Simulator transitions & 7581 & 6538 & 6440 & 6133 \\
\hline
Stage coverage & \textbf{77.56\%} & 61.15\% & 63.67\% & 58.27\% \\
Positive examples & \textbf{3769} & 1946 & 2137 & 1889 \\
\hline
\end{tabular}
\end{table}

\paragraph{Collection efficiency and data yield.}
Table~\ref{tab:collection-budget} contextualizes these results. Under the same seeded task allocation, MAGMA-GEN uses fewer 20B calls than the capacity-privileged baselines, while using 5012 calls to the 4B rollout policy and more simulator transitions. This allocation reaches 77.56\% of the task stages and produces 3769 positive examples, compared with at most 63.67\% stage coverage and 2137 examples for the baselines. Thus, MAGMA-GEN converts cheaper exploration and targeted 20B coaching into broader task coverage rather than relying on more 20B calls. Because generated-token counts were not logged and the methods use different resource mixtures, this comparison does not establish exact compute equivalence.

\begin{table}[h]
\centering
\footnotesize
\caption{Performance with branching factor $B=2$ and real-robot transfer. Simulation metrics are means$\pm$std over four seeds. Real-robot success rates are evaluated on a physical Franka Emika robot using the checkpoint from Seed 1.}
\label{tab:real_robot}
\begin{tabular}{l|c|c|c}
\hline
\textbf{Method} & \textbf{CSR} & \textbf{Recovery} & \textbf{Real Robot (OOD)} \\
\hline
CLEANER-adapted~\cite{cleaner2026} & 15.82{\scriptsize $\pm$ 1.90} & 12.77{\scriptsize $\pm$ 2.08} & 11/30 \\
Expert Rollout Distill. ($B=2$) & 26.99{\scriptsize $\pm$ 2.87} & 18.33{\scriptsize $\pm$ 2.06} & 12/30 \\
\textbf{MAGMA-GEN ($B=2$)} & \textbf{38.35}{\scriptsize $\pm$ 3.97} & \textbf{40.0}{\scriptsize $\pm$ 3.25} & \textbf{15/30} \\
\hline
\end{tabular}
\end{table}

\paragraph{Q3: Scale and Real Robot.}
As summarized in Table~\ref{tab:real_robot}, increasing MAGMA-GEN's branching factor from $B=1$ to $B=2$ raises mean CSR from 16.79\% to 38.35\% and recovery from 26.25\% to 40.0\%. Under the same $B=2$ stage-wise tree-expansion control, Expert Rollout Distillation reaches 26.99\% CSR and 18.33\% recovery. The 20B teacher expands two actions per state using the same subtree-success score as Eq.~\ref{eq:subtree_success}, but without failure-localized interventions; full control details are in Appendix~\ref{app:control}.

On the physical Franka Emika robot, MAGMA-GEN succeeds in 15/30 trials, compared with 12/30 for Expert Rollout Distillation and 11/30 for CLEANER. Under oracle low-level execution, MAGMA-GEN and Expert Rollout Distillation both succeed in $4/6$ trials. With the learned perception and grasping stack, MAGMA-GEN succeeds in $11/24$, compared with $8/24$ and $9/24$, respectively. These latter trials include missed detections, failed grasps or plans, and human object moves, and therefore directly exercise recovery under deployment errors. The frozen policies use a custom segmentation model, M2T2~\cite{m2t2}, and HPP~\cite{hpp}, without a coach, simulator, or digital twin. With one checkpoint and 30 trials per method, these results establish transfer feasibility rather than statistical superiority.

\subsection{Mechanism Ablation}
\label{sec:ablation}

\begin{table}[h]
\centering
\small
\caption{Core mechanism ablation.}
\label{tab:coach_ablation}
\resizebox{\linewidth}{!}{%
\begin{tabular}{l|cc|ccc}
\hline
\textbf{Variant} & \textbf{Coach proposals} & \textbf{Re-exec. valid.}
& \textbf{CSR} & \textbf{Recovery} & \textbf{CGC} \\
\hline
No Coaching & No & No & 4.99{\scriptsize $\pm$ 0.38} & 1.66{\scriptsize $\pm$ 1.44} & 14.57{\scriptsize $\pm$ 1.57} \\
No Re-Execution & Yes & No & 8.21{\scriptsize $\pm$ 0.16} & 5.00{\scriptsize $\pm$ 0.00} & 20.06{\scriptsize $\pm$ 1.50 } \\
\textbf{MAGMA-GEN} & \textbf{Yes} & \textbf{Yes}
& \textbf{16.79}{\scriptsize $\pm$ 2.98} & \textbf{26.25}{\scriptsize $\pm$ 3.75} & \textbf{30.74}{\scriptsize $\pm$ 3.98} \\
\hline
\end{tabular}%
}
\end{table}

\paragraph{Are coaching and re-execution both necessary?}
Table~\ref{tab:coach_ablation} isolates the two core components. Recovery falls from 26.25\% to 5.00\% when coach proposals are retained without re-execution validation, and to 1.66\% without coaching. CSR and CGC follow the same ordering. Without re-execution validation, coach proposals are used as labels without checking whether they improve downstream progress. Results show that both mechanisms are crucial for collecting meaningful data from rollouts by guiding exploration and verifying proposal quality through execution.

\paragraph{Does diagnosis improve correction validity?}
Of 3114 MAGMA-GEN coach proposals (84.8\% corrections and 15.2\% recovery actions), 54.54\% are counterfactually validated, compared with 30.97\% of CLEANER's immediate corrections under the same criterion. This is consistent with diagnosis helping identify the decision responsible for a failure before correction. Coaching also changes which states are reached: 1464 of the 3769 retained examples are selected coach proposals, and 1536 are downstream on-policy states reached through those interventions. Together, they account for 79.60\% of the dataset. All statistics in this paragraph are averaged over the four generation runs.

\paragraph{Does a robust policy require stochastic training failures?}
To measure this effect, we compare MAGMA-GEN with and without injected execution failures during data generation using four runs. Injection improves Recovery from 14.17 to 26.25 and CSR from 12.05 to 16.79, suggesting that stochastic training failures expose the policy to off-nominal states needed for recovery.

%===============================================================================

\section{Limitations}
\label{sec:limitations}

MAGMA-GEN addresses failures whose consequences can be handled through the
available high-level actions. It cannot recover from physical states for which
the existing skills offer no valid continuation. Extending diagnose--propose--validate to continuous control would
likely require a different intervention proposer, such as a visuomotor policy or
trajectory optimizer, together with suitable execution-level validation. During
data generation, MAGMA-GEN also requires privileged access to restorable simulator
states and sufficiently faithful task-level action consequences. Mismatches in
execution or observation dynamics can make validated interventions ineffective
on the physical robot. Extending validation to physical settings where resets
are unavailable or expensive remains an open challenge.

Finally, absolute success rates remain modest because the
benchmark deliberately combines long horizons, evolving rules, partial
observations, and execution-induced failures. MAGMA-GEN improves the supervision
signal, but the underlying language-policy architecture is not designed for
robust long-horizon memory and planning. Developing architectures better matched
to this regime is left to future work.

\section{Conclusion}
\label{sec:conclusion}

We introduced MAGMA-GEN, a data-generation framework that turns ambiguous
on-policy failures into validated recovery supervision for long-horizon
interactive manipulation. At its core is a coaching mechanism that diagnoses
suspected decision-level errors and proposes localized corrections or recovery
actions. Rather than treating every failed outcome as a policy error, MAGMA-GEN
uses matched counterfactual re-execution to filter these fallible diagnoses and
accepts only branches that improve downstream progress. It also provides \emph{guided exploration} by injecting plausible recovery actions into the learner's own failure distribution, reaching
informative off-nominal states that unguided rollouts may rarely discover. The resulting validated branches can be used with standard supervised
fine-tuning, avoiding the sparse rewards and unstable long-horizon credit
assignment often faced by reinforcement learning. Our results indicate that this
simple training interface improves recovery after execution-induced failures and
increases progress before unrecoverable failure, making MAGMA-GEN a practical
path toward more consistent high-level robotic decision making under evolving
goals and structured uncertainty.

%===============================================================================

\clearpage
% The acknowledgments are automatically included only in the final and preprint versions of the paper.
\acknowledgments{We thank Arthur Tanneau, Justin Ganivet, and Benjamin Le Rohellec from Siléane's research and development team, as well as Abdelbasset Houdass, for their contributions to the development of the task environment.}

%===============================================================================

\bibliography{example}  % .bib

\clearpage
\appendix
% Keep appendix routing in one place as new appendix sections are added.
\section{Extended Problem Formalization}
\label{app:problem-formalization}

This appendix refines Section~\ref{formulation} by making the turn chronology explicit and by detailing the kernels.

\subsection{Turn Chronology}
Each discrete time step $t$ represents one cycle of interaction. Given the current state $s^t$ and the agent's action $a^t$, the transition to the next step follows a specific causal sequence:
\begin{enumerate}
    \item \textbf{Physical Execution:} The world transitions to $w^{t+1}$ based on the tool call $a_m^t$.
    \item \textbf{Event Emission:} An interaction event $q^{t+1}$ is generated (either environment feedback or a user instruction).
    \item \textbf{Task Evolution:} The latent task $z^{t+1}$ is updated \textit{conditioned on the new event} $q^{t+1}$.
    \item \textbf{Observation:} The agent receives $o^{t+1} = (k^{t+1}, q^{t+1})$ to begin the next turn.
\end{enumerate}

The first instruction comes at $q^1$, and $w^0, s^0$ are initialized. This chronology covers ordinary execution turns, purely conversational turns with no tool invocation, and interruptions in which a new instruction refines, replaces, or redirects the active task specification.

\subsection{Explicit factorization of the kernels}

The agent first executes a tool, the physical world evolves stochastically, and the next interaction event is generated from the realized post-execution state or by the user. We write this dependency as
\begin{equation}
\begin{aligned}
p(s^{t+1}, o^{t+1} \mid s^t, a^t) = & \underbrace{T_w(w^{t+1} \mid w^t, a_m^t)}_{\text{Physical Dynamics}} \cdot \underbrace{\Omega_q(q^{t+1} \mid w^{t+1}, k^t, a^t)}_{\text{Interaction Event}} \\
& \cdot \underbrace{T_z(z^{t+1} \mid z^t, q^{t+1})}_{\text{Latent Task Update}} \cdot \underbrace{T_k(k^{t+1} \mid k^t, q^{t+1}, a^t)}_{\text{Attribute Update}}
\end{aligned}
\end{equation}
Here, $T_w$ captures stochastic execution, $\Omega_q$ captures user instructions or tool feedback, and $T_z$ updates the latent task state when new instructions modify the goal or active rules. The key point is that tool feedback is generated from the state reached after execution, not directly from the validity of the selected high-level decision.

Because $k^{t+1}$ is observed without noise and $o^{t+1} = (k^{t+1}, q^{t+1})$, writing the joint kernel over $(s^{t+1}, o^{t+1})$ is equivalent to writing a kernel over
$(w^{t+1}, z^{t+1}, k^{t+1}, q^{t+1})$. 

\paragraph{Physical World Dynamics ($T_w$):} 
If no tool is called ($a_m^t = \mathtt{null}$), the physical state remains stationary. Otherwise, it evolves stochastically:
\begin{equation}
    w^{t+1} \sim 
    \begin{cases} 
        T_w(\cdot \mid w^t, a_m^t) & \text{if } a_m^t \neq \mathtt{null} \\
        \delta_{w^t} & \text{if } a_m^t = \mathtt{null}
    \end{cases}
\end{equation}
where $\delta_x$ denotes the point mass at $x$.

\paragraph{Interaction Event Kernel ($\Omega_q$):} 
The event kernel has two branches. If the next event is a new instruction, it is generated by an exogenous instruction process. If the next event is tool feedback, it is generated from the post-execution world state reached by the selected tool:
\begin{equation}
\label{eq:oq-factorization}
    q^{t+1} \sim 
    \begin{cases}
        \Omega_{\mathrm{env}}(\cdot \mid w^{t+1}, a_m^t) & \text{if } a_m^t \neq \mathtt{null} \quad \text{(Tool Feedback)} \\
        \Omega_{\mathrm{instr}}(\cdot \mid w^{t+1}, a_u^t) & \text{if } a_m^t = \mathtt{null} \quad \text{(User Instruction)}
    \end{cases}
\end{equation}
In deployment, $\Omega_{\mathrm{instr}}$ corresponds to the user. During data
generation, this branch is instantiated by SimUser
(Appendix~\ref{app:simuser}). The second branch makes explicit the tool-specific
feedback kernel $\Omega_{\mathrm{env}}$, which depends on the
realized post-execution state $w^{t+1}$, not only on the symbolic tool token.

\paragraph{Latent Task Update ($T_z$):} 
The latent task $z = (g, R_{\mathrm{u}})$ is updated only when the event $q^{t+1}$ is an instruction. Tool feedback informs the agent about the world $w$, but does not change the underlying goal or active rules:
\begin{equation}
\label{eq:tz-factorization}
    z^{t+1} = 
    \begin{cases}
        \Phi(z^t, q^{t+1}) & \text{if } q^{t+1} \in \mathcal{I} \\
        z^t & \text{if } q^{t+1} \in \mathcal{E}
    \end{cases}
\end{equation}
The update function $\Phi$ allows an instruction to simultaneously shift the goal $g$ and refine the active rule set $R_{\mathrm{u}}$.

\paragraph{Structured Attributes ($T_k$):} 
Unlike the latent $z$, the attributes $k$ are system-maintained and fully observed. Their update is a deterministic mapping $\Psi$ based on the latest interaction and action:
\begin{equation}
\label{eq:tk-factorization}
    k^{t+1} = \Psi(k^t, q^{t+1}, a^t)
\end{equation}

\section{Rollout Tree Expansion}
\label{app:exploration}

This appendix complements Sec.~\ref{sec:magma-gen} by detailing how
MAGMA-GEN keeps rollout-tree generation both finite and semantically diverse.
The two relevant mechanisms are stage-wise control of tree expansion and the
simulation of user instructions. Both operate on the data-collection process;
they do not change the task semantics or the success criterion used for
supervision.

\subsection{Stage-Wise Tree Expansion Control}
\label{app:control}

Sec.~\ref{sec:magma-gen} describes MAGMA-GEN as expanding a set of candidate
actions $\mathcal{B}_{\theta}(m^t,o^t)$ from each visited state. Naively
propagating every successful branch through a long-horizon task would make the
number of rollouts grow exponentially. We therefore control expansion at the
stage level. For each stage $j$, successful partial trajectories are first
collected in a stage-local pool, and only a bounded subset is released as
parents for stage $j+1$.

The release rule balances feasibility and coverage. Branches that satisfy the
active goal and rules are eligible for propagation, but branches with nearly
identical ancestor decisions are down-sampled so that one early decision does
not dominate the tree. This keeps the explored subtree close to the policy's
own support while preserving multiple distinct continuations for the same
stage. When later stages do not receive enough viable parents, the rule is
relaxed and additional successful branches are released from upstream pools.
Thus, the generator avoids both uncontrolled branching and premature collapse
to a single trajectory.

This stage-wise allocation also matches the comparison logic of
Eq.~\ref{eq:subtree_success}. Physical and observation perturbations are sampled at the
beginning of a stage and held fixed across sibling branches. Differences in
subtree success therefore reflect the selected action and its downstream
continuation, rather than unrelated resampling of the environment or user
observation process.

\subsection{User Simulation and Semantic Randomization}
\label{app:simuser}

During data generation, the user-instruction branch $\Omega_{\mathrm{instr}}$ from Appendix~\ref{app:problem-formalization} is instantiated by SimUser, implemented with GPT-OSS-20B~\cite{gptoss20}. SimUser generates the next natural-language instruction at stage boundaries, conditioned on the current dialogue. Its role is limited to linguistic realization: the generated instruction must preserve the same latent goal and active rules, without adding new actions or environment assumptions. MAGMA-GEN further randomizes the interface exposed to the policy, including tool names, tool descriptions, parameter names, attribute names, and object surface forms. 

Together, SimUser and interface randomization broaden the observed language and schema distribution while keeping the latent task constant.

\section{Failure Taxonomy}
\label{app:failure}

This appendix details the coarse failure labels used in Sec.~\ref{sec:magma-gen}. They are not treated as supervision: their role is to restrict the space of plausible local interventions and to condition the coach diagnosis.

\subsection{Stage-level failures}
These failures concern the relation between the executed trajectory and the
active latent task $z^t=(g^t,R_{\mathrm{u}}^t)$. A \emph{goal failure}
occurs when the branch does not complete $g^t$ or violates an active rule in
$R_{\mathrm{u}}^t$. A \emph{suboptimal completion} satisfies
the active goal and rules but exceeds the target action budget defined by the
task.

\subsection{Interface-level failures.}

These failures are more localized and usually constrain the intervention space
to a syntactic or communication-level repair. A \emph{missing action} occurs
when the policy emits no tool call although the stage requires one. An
\emph{invalid tool call} occurs when the policy selects a non-existent tool,
uses invalid argument names, violates the tool schema, or refers to objects not
available in the current structured state $k^t$. A \emph{format failure} occurs
when the output cannot be parsed into the expected action format. A
\emph{textual consistency failure} occurs when the user-facing response $a_u^t$
contradicts, ignores, or prematurely confirms the active goal or rules.

In these special cases, the coach diagnosis only produces an explanation, as the faulty step is already identified.

\subsection{Sequential attribution.}
A trajectory can contain several failure modes, so MAGMA-GEN does not assume that repairing the first visible error repairs the trajectory. For example, fixing an invalid argument name can make a tool call executable while revealing that the selected tool is still inappropriate for the active goal. The corrected branch may then fail later, or fail at the same semantic decision for a different reason. The coaching mechanism can therefore apply multiple diagnoses to the same trajectory.

Thus, failure labels condition diagnosis and local correction, but dataset quality is determined by re-execution and subtree validation: a branch contributes supervision only if it yields a more viable continuation under the same stage conditions.

\paragraph{Failure handling and recovery boundaries.}
Decision and interface errors can be addressed by correcting an existing high-level tool call. For example, in Color Sorting, selecting a yellow cube before satisfying an active green-first rule calls for a correction that selects a green cube instead. Missed detections, failed grasps, and reachability failures can instead occur even when the original decision is valid. The decision error may arise afterward: following a failed grasp with a placement command ignores the execution outcome. The coach can localize this subsequent decision and propose another grasp attempt on an eligible cube while preserving the active color-order rule. In our simulated experiments, perturbations are sampled so that at least one valid continuation remains available to complete the subgoal.

MAGMA-GEN cannot recover when no continuation using the available skills can satisfy the subgoal. It does not train new perception, grasping, or control skills (Sec.~\ref{sec:limitations}). Coach proposals may also be ineffective: candidates with zero downstream stage success cannot be selected as supervision under Eq.~\ref{eq:suc}. These examples describe the method's scope; the aggregate results in Sec.~\ref{sec:result} do not establish gains separately for each failure type.

\section{Dynamic Task Building}
\label{app:task-building}

MAGMA-GEN generates training tasks from compact, human-authored task
definitions rather than from a manually enumerated list of demonstrations or
prompts. A task definition specifies the scenario envelope: the simulator
environment, the available tool API, the initial symbolic task state, the
randomization configuration, and a set of admissible user-request generators.
The symbolic state contains the task-level facts that should persist across an
episode, such as known objects, target areas, object--area assignments,
categories, temporary prohibitions, and memory.

\subsection{Generator}

Each request generator is a state-conditioned task primitive. Given the current
symbolic state, it exposes: (i) a sampling weight, which is zero when the
primitive is not currently admissible; (ii) a stage constructor, which produces
one or more executable task stages, including the user instruction, required
goals, and target step budget; and (iii) an optional state update, which records
the consequences of the sampled request for subsequent stages. Thus, random
generation is not an independent draw of text templates: it is a stochastic
composition of valid task transitions.

At the beginning of generation, the task generator builds a default task from
the selected definition and clones its initial state. It then repeatedly
evaluates all active request generators on the current state, discards the
generators with zero weight, samples one remaining generator proportionally to
its weight, appends the generated stage or stages, and applies the corresponding
state update. This loop stops when no request remains admissible or when the
maximum target-step budget is reached. The resulting task is therefore a
coherent multi-stage episode whose later instructions depend on earlier sampled
events.

\subsection{Example}

In the warehouse-sorting scenario, for example, the definition provides the
warehouse environment, the sorting tools, the initial set of objects and target
areas, and the active request families. These families include adding or
removing known areas, assigning objects or categories to areas, launching
sorting cycles, introducing temporary forbidden objects, and issuing one-shot
cycle overrides. The generator may therefore sample sequences such as: first
introduce a new default assignment, then request a sorting cycle that uses this
assignment, then temporarily forbid an object and sample a later instruction
that must be rejected because it violates the current rule. Such dependencies
come from the evolving symbolic state, not from hand-written episode scripts.

\subsection{Human role}

The human role is consequently concentrated in scenario design. A developer
declares the initial state, the admissible request primitives, their
natural-language realizations, their validity conditions, and the goals used to verify
execution. MAGMA-GEN then performs the per-episode composition automatically,
yielding many training tasks from the same compact specification. Humans can
still declare direct presets when an exact task is needed for debugging,
evaluation, or ablation, but the standard training data are produced by this
dynamic generator with low per-task manual effort.

\section{Experimental Protocol}
\label{app:protocol}

\subsection{Baseline Implementations}

All baselines are integrated into the ManiSkill3 simulator \cite{taomaniskill3} and use the same simulator settings, code environment, task budget, evaluation protocol, stochastic execution, and randomization settings (see Appendix~\ref{app:exploration}) as our method. Thus, baselines are not collected from clean rollouts: grasp failures, reachability failures, partial detections, and tool-feedback perturbations are sampled from the same execution and observation distributions during data generation. The correction modules of MAGMA-GEN, CLEANER-adapted, and AutoTraj-adapted receive the same privileged task specification and simulator-derived failure category, and use the same 20B corrector and simulator interface. Their trajectory context depends on when correction occurs: CLEANER sees the executed prefix up to the detected error, MAGMA-GEN sees the completed failed subtrajectory, and AutoTraj sees the completed failed expert trajectory. Thus, they share the same categories of privileged information, while differing in the temporal context available for correction, how they locate the responsible action, and whether the proposal is validated from a restored state. Each method produces offline supervision data that are then used to train the same base policy, Qwen3-4B \cite{qwen3technicalreport}, with the supervised fine-tuning protocol described below.

\paragraph{Expert Rollout Distillation.}
This baseline is implemented with gpt-oss:20b \cite{gptoss20}. The model is prompted with the privileged task specification and chain-of-thought examples to improve its reliability on the different subgoals; it does not receive failed trajectories as correction input. We keep only successful rollouts collected under the same stochastic execution process and use them as demonstrations for supervised fine-tuning. This baseline tests whether a stronger teacher, without explicit recovery supervision, is sufficient to train the smaller policy. For the $B=2$ variant, reported as Expert Rollout Distillation ($B=2$), we generate two candidate actions per expanded state. We use the same tree-expansion control mechanism as in our method (see Appendix~\ref{app:control}) and select supervision with the same subtree-success score.

\paragraph{Best-of-N Distillation.}
This baseline extends rollout distillation by allowing the collector to retry a failed subgoal up to $N$ times. We use $N=10$: when a subgoal fails, generation restarts from the beginning of that subgoal, preserving the successful prefix, until either a successful continuation is found or the retry budget is exhausted. We report both an on-policy variant, where Qwen3-4B performs the retries, and an expert variant, where gpt-oss:20b performs the retries. Successful retry branches are retained as supervised demonstrations. There is no branching factor because the baseline creates retry branches directly.

\paragraph{CLEANER-adapted.}
We follow the concept of CLEANER \cite{cleaner2026}, but adapt it to data collection in our robot-command setting. During simulation, when the first detectable error is observed among format errors, invalid calls, missing actions, or textual failures (see Appendix~\ref{app:failure}), the correction module immediately replaces the action at which the error becomes visible and execution continues. If the corrected continuation still does not complete the subgoal, we relaunch from the beginning of that subgoal while preserving the successful prefix, for at most 10 attempts. CLEANER therefore measures whether online detected-step replacement is sufficient. Unlike MAGMA-GEN, it does not inspect the completed failed subtrajectory to attribute the failure to an earlier decision or distinguish such a decision error from a valid action followed by an execution failure.

\paragraph{AutoTraj-adapted.}
We follow the concept of AutoTraj \cite{autotraj} by taking a failed trajectory from the Expert Rollout Distillation collector and prompting gpt-oss:20b to produce a complete corrected trajectory in one pass. The correction is performed post hoc, after the failed rollout is complete, and is added as offline supervision without restoring the simulator state or re-executing the proposed repair for validation. Our adaptation focuses on AutoTraj's trajectory-repair component in the robot-command setting; it omits the original trajectory reward model and subsequent reinforcement-learning stage to keep the training objective identical across methods. All repaired data instead use the same SFT protocol as the other methods. This baseline therefore measures whether complete-trajectory repair alone can convert failed expert rollouts into useful supervision.

\paragraph{Difference from MAGMA-GEN.}
MAGMA-GEN also waits until a subtrajectory has failed, but uses the complete failed subtrajectory to diagnose which earlier decision most likely made it non-viable. The coach then proposes either a replacement at that localized decision or a recovery action after an execution failure. The simulator is restored to the corresponding parent state, and the intervention is retained only when matched re-execution improves downstream completion. This separates causal localization and execution-based validation from CLEANER's immediate detected-step replacement and AutoTraj's unvalidated post-hoc repair.

\subsection{Detailed Protocol}

All methods are run on the same fixed collection plan. Within each of the four generation runs, they receive the exact same 180 seeded task instances, including the same task definitions, ordered subgoals, and target-step limits. Best-of-N and CLEANER are capped at $N=10$ retries or correction-and-relaunch attempts per subgoal. AutoTraj instead applies one complete post-hoc repair to each failed Expert Rollout Distillation trajectory, while MAGMA-GEN allocates its attempts to coached counterfactual continuations. A method that cannot complete a subgoal within its collection procedure does not receive the later subgoals of that task. Consequently, stage coverage and final dataset size measure how far each mechanism progresses through the common task exposure.

\begin{center}
\begin{tabular}{lcc}
\hline
Task block & Number of tasks & Target-step cap \\
\hline
Warehouse sorting, simple & 7 & 12 \\
Warehouse sorting, forbidden & 8 & 12 \\
Warehouse sorting, category & 15 & 12 \\
Coffee making & 15 & 20 \\
Coffee making, team setting & 15 & 20 \\
Delivery & 30 & 12 \\
Color sorting & 25 & 20 \\
Color sorting, clean-table preset & 5 & 15 \\
Laundry & 30 & 20 \\
Press buttons & 30 & 20 \\
\hline
Total & 180 & -- \\
\hline
\end{tabular}
\end{center}

We fix the task budget rather than the final number of supervised examples. Consequently, methods with stronger supervision can produce more successful trajectories and therefore larger SFT datasets.

All policies are fine-tuned from Qwen3-4B with LoRA adapters only. We use rank $r=32$, $\alpha=64$, dropout $0.1$, no bias parameters, and apply LoRA to the \texttt{q\_proj}, \texttt{k\_proj}, \texttt{v\_proj}, \texttt{o\_proj}, \texttt{gate\_proj}, \texttt{up\_proj}, and \texttt{down\_proj} modules. Training is performed for 3 epochs with per-device batch size 4 and gradient accumulation over 8 steps, giving an effective batch size of 32 on our single-GPU setup. We use AdamW, a learning rate of $10^{-5}$, cosine learning-rate decay, warmup ratio $0.05$, weight decay $0.01$, and bf16 precision. The maximum sequence length is 3000 tokens. Training is run on one 48GB NVIDIA Ada GPU without DeepSpeed.

Generated positive trajectories are not held out for validation; all are used for training. We use an external validation set for monitoring, and we select the best checkpoint at 0.5-epoch intervals.

\section{Scenario Definitions and Evaluation Composition}
\label{app:scenario-description}

Appendix~\ref{app:task-building} describes how training episodes are sampled from compact task definitions. For evaluation, we instead use human-authored benchmark tasks with fixed seeds, fixed perturbation schedules, and explicit success oracles. Table~\ref{tab:scenario-summary} summarizes the role of each scenario in the benchmark. Each task instantiates a small number of rule families and checks whether the agent preserves the resulting latent task state across later instructions, tool feedback, and temporary overrides.

\begin{table}[t]
\centering
\footnotesize
\setlength{\tabcolsep}{3pt}
\caption{Scenario summary. The Regime column indicates each scenario's role in the benchmark; scenarios marked with \textsuperscript{*} are evaluated on the real robot. Each scenario is expanded below with its tools, rule families, and one representative evaluation task.}
\label{tab:scenario-summary}
\begin{tabular}{p{0.18\linewidth}p{0.12\linewidth}p{0.39\linewidth}p{0.20\linewidth}}
\hline
\textbf{Scenario} & \textbf{Regime} & \textbf{Primary capability} &
\textbf{Uncertainty} \\
\hline
Warehouse Sorting &
In-domain &
Persistent object--area assignments, category rules, and one-shot exceptions. &
None. \\

Make Coffee &
In-domain &
Preference memory over people and teams, plus registry queries. &
Unreachable pods. \\

Color Sorting &
In-domain &
Grounding color-order rules into object-level manipulation. &
Failed grasps, unreachable cubes, and partial detections. \\

Sorting &
OOD \textsubscript{*} &
Storing objects in containers. &
Failed grasps and partial detections. \\

Laundry &
Comp. &
Transfer from direct object assignments to category-to-detergent rules. &
Unreachable clothes. \\

Delivery &
Training only &
Recipe memory, recipe deltas, order metadata, and one-shot recipe overrides. &
Partial detection and grasp failures. \\

Press Button &
Training only &
Ordering rules, prefix rules, exact-order requests, and selective forgetting. &
None. \\

\hline
\end{tabular}
\end{table}

\paragraph{Common task structure.}
Each task (training or evaluation) is a multi-goal dialogue. A subgoal can:
\begin{itemize}
    \item Update persistent rules, such as default assignments or user preferences
    \item Request an action under the current rules
    \item Ask about the current state
\end{itemize}
A subgoal is successful when the agent emits the expected tool call sequence or user-facing answer while respecting all active rules. Under injected uncertainty, execution feedback can report a failed grasp, missed detection, unreachable object, or failed physical action. The agent is then expected to recover when a valid alternative exists, and to refuse or ask for clarification when no valid continuation satisfies the active task state.

\subsection{Warehouse Sorting}

Warehouse Sorting is the canonical symbolic-memory scenario. The agent controls a
sorting cell by launching complete cycles from object--area assignments rather
than manipulating objects individually.

The real-robot sorting evaluation uses a held-out physical sorting variant and
is counted as OOD in Table~\ref{tab:scenario-summary}.

\textbf{Available functions:}
\begin{enumerate}
    \item \textit{launch\_cycle(assignment : Dict)}
    \item \textit{add\_area(name : str)}
    \item \textit{remove\_area(name : str)}
\end{enumerate}

\textbf{Rule families:}
\begin{enumerate}
    \item Assign one object, a set of objects, or a category to an area.
    \item Mark an object as forbidden for future cycles.
    \item Express equality rules, e.g., object X follows object Y.
    \item Add or remove target areas during the dialogue.
    \item Apply a one-shot exception without overwriting the default assignment.
\end{enumerate}

\paragraph{Representative evaluation task.}
Table~\ref{tab:warehouse-example-task} shows one Warehouse Sorting task used to
check whether the agent updates persistent assignments while keeping one-shot
exceptions local to the current cycle.

\begin{table}[ht]
\centering
\small
\caption{Representative Warehouse Sorting evaluation task. The task combines global rules, persistent updates, and one-shot exceptions.}
\label{tab:warehouse-example-task}
\begin{tabular}{p{0.05\linewidth} p{0.43\linewidth} p{0.42\linewidth}}
\hline
\textbf{Step} & \textbf{User instruction} & \textbf{Expected task state / action} \\
\hline
1 &
Today, the sorting rule is: all objects to \texttt{area1}. &
Store default rule: all objects $\rightarrow$ \texttt{area1}. \\

2 &
Launch a cycle. &
Execute: $\{\texttt{ref\_obj\_*} \rightarrow \texttt{area1}\}$. \\

3 &
For all future cycles, \texttt{ref\_obj\_1} must go to \texttt{area2}. &
Update default: \texttt{ref\_obj\_1} $\rightarrow$ \texttt{area2}. \\

4 &
Launch a cycle for all objects. &
Execute: $\{\texttt{ref\_obj\_2}, \texttt{ref\_obj\_3} \rightarrow \texttt{area1};
\texttt{ref\_obj\_1} \rightarrow \texttt{area2}\}$. \\

5 &
Launch a cycle according to the default assignment of each object. Except for
\texttt{ref\_obj\_3}, send it to \texttt{area2}. &
Execute temporary exception:
$\{\texttt{ref\_obj\_2} \rightarrow \texttt{area1};
\texttt{ref\_obj\_1}, \texttt{ref\_obj\_3} \rightarrow \texttt{area2}\}$. \\

6 &
The default target for \texttt{ref\_obj\_2} is now \texttt{area3}. &
Update default: \texttt{ref\_obj\_2} $\rightarrow$ \texttt{area3}. \\

7 &
Launch a cycle for all objects. &
Execute:
$\{\texttt{ref\_obj\_3} \rightarrow \texttt{area1};
\texttt{ref\_obj\_1} \rightarrow \texttt{area2};
\texttt{ref\_obj\_2} \rightarrow \texttt{area3}\}$. \\
\hline
\end{tabular}
\end{table}

\subsection{Make Coffee}

Make Coffee combines remembered preferences with multi-step tool use and
external information retrieval. The same request can be underspecified because a
person's coffee type may be defined directly or inherited from a team.

\textbf{Available functions:}
\begin{enumerate}
    \item \textit{start\_machine()}
    \item \textit{load\_capsule(pod : str)}
    \item \textit{place\_mug()}
    \item \textit{get\_team\_name(name : str)}
    \item \textit{get\_team\_members(team\_name : str)}
\end{enumerate}

\textbf{Rule families:}
\begin{enumerate}
    \item Store individual coffee preferences.
    \item Store team-level preferences and apply them to team members.
    \item Query the registry to resolve a person's team or list team members.
    \item Serve several users while preserving the required preparation sequence.
\end{enumerate}

\paragraph{Representative evaluation task.}
Table~\ref{tab:make-coffee-example-task} targets multi-step tool use and
information retrieval. The agent must remember a team-level coffee preference,
serve multiple users with the right preparation sequence, and later answer a
registry question.

\begin{table}[ht]
\centering
\small
\caption{Representative Make Coffee evaluation task. The task combines persistent team-level preferences, repeated multi-step execution, and registry-based question answering.}
\label{tab:make-coffee-example-task}
\begin{tabular}{p{0.06\linewidth} p{0.42\linewidth} p{0.42\linewidth}}
\hline
\textbf{Step} & \textbf{User instruction} & \textbf{Expected task state / action} \\
\hline
1 &
DISCO team only drinks black coffee. &
Store preference: members of DISCO $\rightarrow$ black coffee. \\

2-8 &
Serve Smith and Clark. &
For each requested person, execute the black-coffee sequence:
\textit{place\_mug()}, \textit{load\_capsule(black)}, then
\textit{start\_machine()}. \\

9-10 &
In which teams are Lee and Scott? &
Use the registry tools and answer that Lee and Scott are both in MAC. \\
\hline
\end{tabular}
\end{table}

\subsection{Color Sorting}

Color Sorting evaluates whether symbolic ordering rules can be grounded into
object-level manipulation under partial observability. The scene contains two
colored boxes and several cubes per color; object identities and availability
must be obtained through perception.

\textbf{Available functions:}
\begin{enumerate}
    \item \textit{get\_object\_state()}
    \item \textit{take\_object\_per\_id(name : str)}
    \item \textit{put\_to\_box(color : str)}
\end{enumerate}

\textbf{Rule families:}
\begin{enumerate}
    \item Sort all cubes of one color before cubes of another color.
    \item Reverse the color priority.
    \item Alternate colors when both colors are requested.
\end{enumerate}

\paragraph{Evaluation and uncertainty.}
The environment contains two boxes and three cubes per color. Perturbations
include failed grasps, unreachable cubes, and partial detections.

\paragraph{Representative evaluation task.}
Table~\ref{tab:color-sorting-example-task} checks whether the agent preserves an
ordering rule while recovering from inaccessible or masked cubes.

\begin{table}[ht]
\centering
\small
\caption{Representative Color Sorting evaluation task. The task combines persistent ordering, object-level grounding, and recovery from failed grasps and masked cubes.}
\label{tab:color-sorting-example-task}
\begin{tabular}{p{0.06\linewidth} p{0.42\linewidth} p{0.42\linewidth}}
\hline
\textbf{Step} & \textbf{User instruction} & \textbf{Expected task state / action} \\
\hline
1 &
For this table, sort green cubes before yellow cubes. &
Store persistent ordering rule: green cubes must be sorted before yellow cubes. \\

2-7 &
Store two yellow cubes and one green cube. &
First place one green cube in \texttt{green\_box\_pose}. If a green cube cannot
be grasped, retry with another green cube when available; do not place a yellow
cube before satisfying the green-cube requirement. Then store two yellow cubes. \\
\hline
\end{tabular}
\end{table}

\subsection{Laundry}

Laundry is a compositional-generalization scenario. Training tasks use direct
object-to-detergent assignments, while evaluation introduces category-level
rules such as white clothes, colored clothes, or everyday clothes. This tests
whether a rule structure learned in sorting-like domains transfers to a
different manipulation domain.

\textbf{Available functions:}
\begin{enumerate}
    \item \textit{take(name : str)}
    \item \textit{drop()}
    \item \textit{action\_wash()}
\end{enumerate}

\textbf{Rule families:}
\begin{enumerate}
    \item Assign clothes or clothing categories to a detergent.
    \item Correct a previous category assignment.
    \item Override the remembered detergent for a single wash.
    \item Reject a wash when requested clothes require incompatible detergents.
\end{enumerate}

\paragraph{Evaluation and uncertainty.}
The environment contains nine clothes: \texttt{shirt}, \texttt{blouse}, \texttt{jeans}, \texttt{pants}, \texttt{short}, \texttt{socks}, \texttt{boxer}, \texttt{panties}, \texttt{cap}, and three detergents: \texttt{detergent\_OMO}, \texttt{detergent\_PERSIL}, and \texttt{detergent\_SKIP}. Clothes can be unreachable, forcing the agent to adapt its behavior.

\paragraph{Representative evaluation task.}
Table~\ref{tab:laundry-example-task} shows a compositional task where the agent
must apply category rules that were not used in training, then update them after
a correction.

\begin{table}[ht]
\centering
\small
\caption{Representative Laundry evaluation task. The task combines category rules, memory update, and multi-step loading.}
\label{tab:laundry-example-task}
\begin{tabular}{p{0.06\linewidth} p{0.42\linewidth} p{0.42\linewidth}}
\hline
\textbf{Step} & \textbf{User instruction} & \textbf{Expected task state / action} \\
\hline
1 &
Rule: white clothes (\texttt{shirt}, \texttt{socks}, \texttt{panties}) use OMO.
Colored clothes (\texttt{jeans}, \texttt{blouse}, \texttt{cap}) use PERSIL.
The rest use SKIP. &
Store category-to-detergent mapping. \\

2-9 &
Wash the white clothes. &
Load \texttt{shirt}, \texttt{socks}, \texttt{panties}, and
\texttt{detergent\_OMO}; call \textit{action\_wash()}. \\

10 &
Actually, \texttt{cap} and \texttt{blouse} will use SKIP now. &
Update the internal mapping. \\

11-15 &
Wash all clothes that use PERSIL. &
Load \texttt{jeans} and \texttt{detergent\_PERSIL}; call
\textit{action\_wash()}. \\
\hline
\end{tabular}
\end{table}

\subsection{Delivery}

Delivery abstracts a production cell where the robot launches packing cycles
from a recipe. It is deliberately tool-simple: the challenge is to preserve the
default recipe while handling manufacturing-order metadata, recipe deltas, and
explicit one-shot recipes.

\textbf{Available function:}
\begin{enumerate}
    \item \textit{launch\_cycle(manufacturing\_order : str, recipe : List[str], delivery\_number : Optional[int])}
\end{enumerate}

\textbf{Rule families:}
\begin{enumerate}
    \item Define or query the default recipe.
    \item Add products to the default recipe or remove products from it.
    \item Launch a default cycle with a manufacturing order and delivery count.
    \item Launch an explicit one-shot recipe without overwriting the default.
\end{enumerate}

\paragraph{Evaluation and uncertainty.}
Delivery is used during training but is not part of the held-out evaluation
reported in Table~\ref{tab:scenario-summary}.

\subsection{Press Button}

Press Button isolates sequencing. It is used only for training.

\textbf{Available function:}
\begin{enumerate}
    \item \textit{press\_button(id : str)}
\end{enumerate}

\textbf{Rule families:}
\begin{enumerate}
    \item Press one button or group before another whenever both are requested.
    \item Press even buttons before odd buttons, or the reverse.
    \item Always prepend a prefix button before requested sequences.
    \item Forget all ordering rules or selectively forget one rule family.
\end{enumerate}

\subsection{Sorting}

Sorting is an out-of-distribution scenario. It combines symbolic order semantics with concrete manipulation: the robot must detect objects on a table, follow a recipe, and put objects into containers according to rules. It combines elements of warehouse sorting, delivery, and color sorting.

\textbf{Available functions:}
\begin{enumerate}
    \item \textit{detect()}
    \item \textit{take(name : str)}
    \item \textit{put(target : str)}
\end{enumerate}

\textbf{Rule families:}
\begin{enumerate}
    \item Remember a default recipe.
    \item Remember object--area assignment.
\end{enumerate}

\paragraph{Evaluation and uncertainty.}
Perturbations include missed detections and grasp failures. A human adds and removes objects on the table while the robot is manipulating.

\section{Supplementary Analysis of Results}
\label{app:result-discussion}

This section clarifies how to interpret the simulated and real-robot results in Sec.~\ref{sec:result}. The main point is that CSR, recovery rate, and CGC do not measure the same failure mode. CSR requires a complete long-horizon episode to succeed, recovery isolates perturbed subgoals after execution feedback, and CGC measures how much of the task is completed before the first unrecoverable failure. 

\subsection{Why does absolute CSR remain low?}

The CSR values in Table~\ref{tab:main_multiseed} should be read as complete episode success under compounding task and execution constraints. A successful episode requires the policy to remember interaction-derived rules, apply temporary exceptions only when appropriate, choose valid high-level tool calls, and react correctly to stochastic execution feedback. 

This is also why the recovery gains are larger than the CSR gains. In Table~\ref{tab:main_multiseed}, MAGMA-GEN improves recovery from $16.66\%$ for the strongest retry baseline to $26.25\%$, while CSR improves more modestly from $14.15\%$ to $16.79\%$. This gap is expected: MAGMA-GEN improves the data used to train recovery behavior, but the deployed policy is still the
same Qwen3-4B backbone with no explicit long-horizon memory module or planner. 

The $B=2$ results in Table~\ref{tab:real_robot} show that CSR can increase substantially when the same backbone is trained with denser validated supervision. This suggests that low absolute CSR is not only a model-size issue: coverage of the generated supervision also matters. At the same time, the remaining errors indicate that data generation alone does not fully address long-horizon state tracking.

\subsection{Why are real-robot differences smaller than simulation differences?}

The real-robot evaluation in Table~\ref{tab:real_robot} is an
out-of-distribution transfer test. The policies are evaluated on six physical tasks with five variations each, using a real perception, grasping, and planning stack. These trials include missed or incorrect detections, failed grasps or plans, and human
object moves. 

As a result, real-robot failures often come from long sequences of perception and manipulation errors that accumulate until the high-level policy loses the active task state or reaches a state where recovery is no longer possible. This compresses the difference between methods: even a better recovery policy can
fail when the low-level stack repeatedly produces inconsistent observations or unrecoverable physical states. Under this harder OOD setting, MAGMA-GEN achieves the highest observed real-robot success rate: $15/30$, compared with $12/30$ for Expert Rollout Distillation and $11/30$ for CLEANER-adapted. Given the limited number of trials and the use of a single checkpoint per method, these results demonstrate transfer feasibility but do not establish statistical superiority.

\end{document}